\documentclass[10pt,twocolumn,letterpaper]{article}

\usepackage{cvpr}
\usepackage{times}
\usepackage{epsfig}
\usepackage{graphicx}
\usepackage{amsmath}
\usepackage{amssymb}

\usepackage{comment}

\usepackage[pagebackref=true,breaklinks=true,letterpaper=true,colorlinks,bookmarks=false]{hyperref}

\cvprfinalcopy 

\def\cvprPaperID{4562} 

\ifcvprfinal\fi
\begin{document}

\title{Attention-based Adaptive Selection of Operations for Image Restoration\\ in the Presence of Unknown Combined Distortions}

\author{
    Masanori Suganuma${}^\text{1,2}$
    \quad
    Xing Liu${}^\text{1}$
    \quad
    Takayuki Okatani${}^\text{1,2}$\\
    ${}^\text{1}$Graduate School of Information Sciences, Tohoku University
    \quad
    ${}^\text{2}$RIKEN Center for AIP\\
    {\tt\small \{suganuma,ryu,okatani\}@vision.is.tohoku.ac.jp}
}

\maketitle
\thispagestyle{empty}

\begin{abstract}
Many studies have been conducted so far on image restoration, the problem of restoring a clean image from its distorted
version. There are many different types of distortion which affect image quality. Previous studies have focused on single types of distortion, proposing methods for removing them.
However, image quality degrades due to multiple factors in the real world. Thus, depending on applications, e.g., vision for autonomous cars or surveillance cameras, we need to be able to deal with multiple combined distortions with unknown mixture ratios. 
For this purpose, we propose a simple yet effective layer architecture of neural networks.
It performs multiple operations in parallel, which are weighted by an attention mechanism to enable selection of proper operations depending on
the input. The layer can be stacked to form a deep network, which is differentiable and thus can be trained in an end-to-end fashion by gradient descent.  The experimental results show that the proposed method works better than previous methods by a good margin on tasks of restoring images with multiple combined distortions.
\end{abstract}

\section{Introduction}
\label{sec:intro}

The problem of image restoration, which is to restore a clean image from its degraded version, has a long history of research. 
Previously, researchers tackled the problem by modeling (clean) natural images, where they design image prior, such as edge statistics \cite{fattal,perrone_total_2014} and sparse representation \cite{ksvd,yang_image_2010}, 
based on statistics or 
physics-based models of natural images. 
Recently, learning-based methods using convolutional neural networks (CNNs) \cite{lecun_gradient_1998,krizhevsky_imagenet_2012} have been shown to work better than previous methods that are based on the hand-crafted priors, and have raised the level of performance on various image restoration tasks, such as denoising \cite{xie_image_2012,zhang_beyond_2017,mem,zhang2018ffdnet}, deblurring \cite{nah2017deep,sun2015learning,Kupyn_2018_CVPR}, and super-resolution \cite{dong2014learning,ledig2016photo,zhang2018image}.

There are many types of image distortion, such as Gaussian/salt-and-pepper/shot noises, defocus/motion blur, compression artifacts, haze, raindrops, etc. Then, there are two application scenarios for image restoration methods.
One is the scenario where the user knows what image distortion he/she wants to remove; an example is a deblurring filter tool implemented in a photo editing software.
The other is the scenario where the user does {\em not} know what distortion(s) the image undergoes but wants to improve its quality, \eg, applications to vision for autonomous cars and surveillance cameras.

In this paper, we consider the latter application scenario. Most of the existing studies are targeted at the former scenario, and they cannot be directly applied to the latter.
Considering that real-world images often suffer from a combination of different types of distortion, we need image restoration methods that can deal with combined distortions with unknown mixture ratios and strengths.

There are few works dealing with this problem. A notable exception is the work of Yu et al.~\cite{Yu_2018_CVPR}, 
which proposes a framework in which multiple light-weight CNNs are trained for different image distortions and are adaptively applied to input images by a mechanism learned by deep reinforcement learning. 
Although their method is shown to be effective, we think there is room for improvements. 
One is its limited accuracy; the accuracy improvement gained by their method is not so large, as compared with application of existing methods for a single type of distortion to images with combined distortions. Another is its inefficiency; it uses multiple distortion-specific CNNs in parallel, each of which also needs pretraining.

In this paper, we show that a simple attention mechanism can better handle aforementioned combined image distortions. 
We design a layer that performs many operations in parallel, such as convolution and pooling with different parameters. We equip the layer with an attention mechanism that produces weights on these operations, intending to make the attention mechanism to work as a switcher of these operations in the layer. 
Given an input feature map, the proposed layer first generates  attention weights on the multiple operations.
The outputs of the operations are multiplied with the attention weights and then concatenated, forming the output of this layer to be transferred to the next layer.

We call the layer {\em operation-wise attention} layer. This layer can be stacked to form a deep structure, which can be trained in an end-to-end manner by gradient descent; hence, any special technique is not necessary for training.
We evaluate the effectiveness of our approach through several experiments. 

The contribution of this study is summarized as follows:
\begin{itemize}
    \item We show that a simple attention mechanism is effective for image restoration in the presence of multiple combined distortions; our method achieves the state-of-the-art performance for the task.
    \item Owing to its simplicity, the proposed network is more efficient than previous methods. Moreover, it is fully differentiable, and thus it can be trained in an end-to-end fashion by a stochastic gradient descent.
    \item We analyze how the attention mechanism behaves for different inputs with different distortion types and strengths. To do this, we visualize the attention weights generated conditioned on the input signal to the layer. 
\end{itemize}

\section{Related Work}

This section briefly reviews previous studies on image restoration using deep neural networks as well as attention mechanisms used for computer vision tasks.

\subsection{Deep Learning for Image Restoration}

CNNs have proved their effectiveness on various image restoration tasks, such as denoising \cite{xie_image_2012,zhang_beyond_2017,mem,zhang2018ffdnet,suganumaICML2018}, deblurring \cite{nah2017deep,sun2015learning,Kupyn_2018_CVPR}, single image super-resolution \cite{dong2014learning,ledig2016photo,haris2018deep,zhang2018learning,zhang2018residual,zhang2018image}, JPEG artifacts removal \cite{dong2015compression,guo2016building,zhang_beyond_2017}, raindrop removal \cite{Qian_2018_CVPR}, deraining \cite{yang2017deep,acmm,derain_zhang_2018,li2018recurrent}, and image inpainting \cite{pathak2016context,IizukaSIGGRAPH2017,semantic,Liu_2018_ECCV}.
Researchers have studied these problems basically from two directions. One is to design new network architectures and the other is to develop novel training methods or loss functions, such as the employment of adversarial training \cite{gan}.

Mao \etal \cite{red} proposed an architecture consisting of a series of symmetric convolutional and deconvolutional layers, with skip connections \cite{srivastava_2015,res} for the tasks of denoising and super-resolution.
Tai \etal \cite{mem} proposed the memory network having 80 layers consisting of a lot of recursive units and gate units, and applied it to denoising, super-resolution, and JPEG artifacts reduction.
Li \etal \cite{li2018recurrent} proposed a novel network based on convolutional and recurrent networks for deraining, i.e., a task of removing rain-streaks from an image. 

A recent trend is to use generative adversarial networks (GANs), where two networks are trained in an adversarial fashion; a generator  is trained to perform image restoration, and a discriminator  is trained to distinguish whether its input is a clean image or a restored one.
Kupyn \etal \cite{Kupyn_2018_CVPR} employed GANs  for blind motion deblurring.
Qian \etal \cite{Qian_2018_CVPR} introduced an attention mechanism into the framework of GANs and achieved the state-of-the-art performance on the task of raindrop removal.
Pathak \etal \cite{pathak2016context} and Iizuka \etal \cite{IizukaSIGGRAPH2017} employed GANs for image inpainting.
Other researchers proposed new loss functions, such as the perceptual loss  \cite{Johnson2016Perceptual,ledig2016photo,Liu_2018_ECCV}.
We point out that, except for the work of \cite{Yu_2018_CVPR} mentioned in Sec.~\ref{sec:intro}, there is no study dealing with combined distortions.

\subsection{Attention Mechanisms for Vision Tasks}

Attention has been playing an important role in the solutions to many computer vision problems \cite{wang2017residual,li2018tell,Hu_2018_CVPR,anderson2018bottom}. While several types of attention mechanisms are used for language tasks \cite{vaswani2017attention} and vision-language tasks \cite{anderson2018bottom,Nguyen_2018_CVPR}, in the case of feed-forward CNNs applied to computer vision tasks,  
researchers have employed the same type of attention mechanism, which generates attention weights from the input (or features extracted from it) and then apply them on some feature maps also generated from the input.
There are many applications of this attention mechanism. Hu \etal \cite{Hu_2018_CVPR} proposed the squeeze-and-excitation block to 
weight outputs of a convolution layer in its output channel dimension, 
showing its effectiveness on image classification.
Zhang \etal \cite{zhang2018image} incorporated an  attention mechanism into residual networks, showing that channel-wise attention contributes to accuracy improvement for image super-resolution.
There are many other studies employing the same type of attention mechanism; \cite{park2018bam,woo2018cbam,Liu_2018_CVPR} to name a few.

Our method follows basically the same approach, but differs from  previous studies in that we consider attention over multiple different operations, not over image regions or channels, aiming at selecting the operations according to the input. To the authors' knowledge, there is no study employing exactly the same approach. Although it is categorized as neural architecture search (NAS) methods, the study of 
Liu \etal \cite{liu2018darts} is similar to ours in that they attempt to select multiple operations by computing weights on them. However, in their method, the weights are fixed parameters to be learned along with network weights in training. The weights are binarized after optimization, and do not vary depending on the input. The study of Veit \etal \cite{veit2018convolutional} is also similar to ours in that operation(s) in a network can vary depending on the input. However, their method only chooses whether to perform convolution or not in residual blocks \cite{res} using a gate function; it does not select mulitple different operations.

\section{Operation-wise Attention Network}

\begin{figure}[t]
\begin{center}
\includegraphics[scale=0.1]{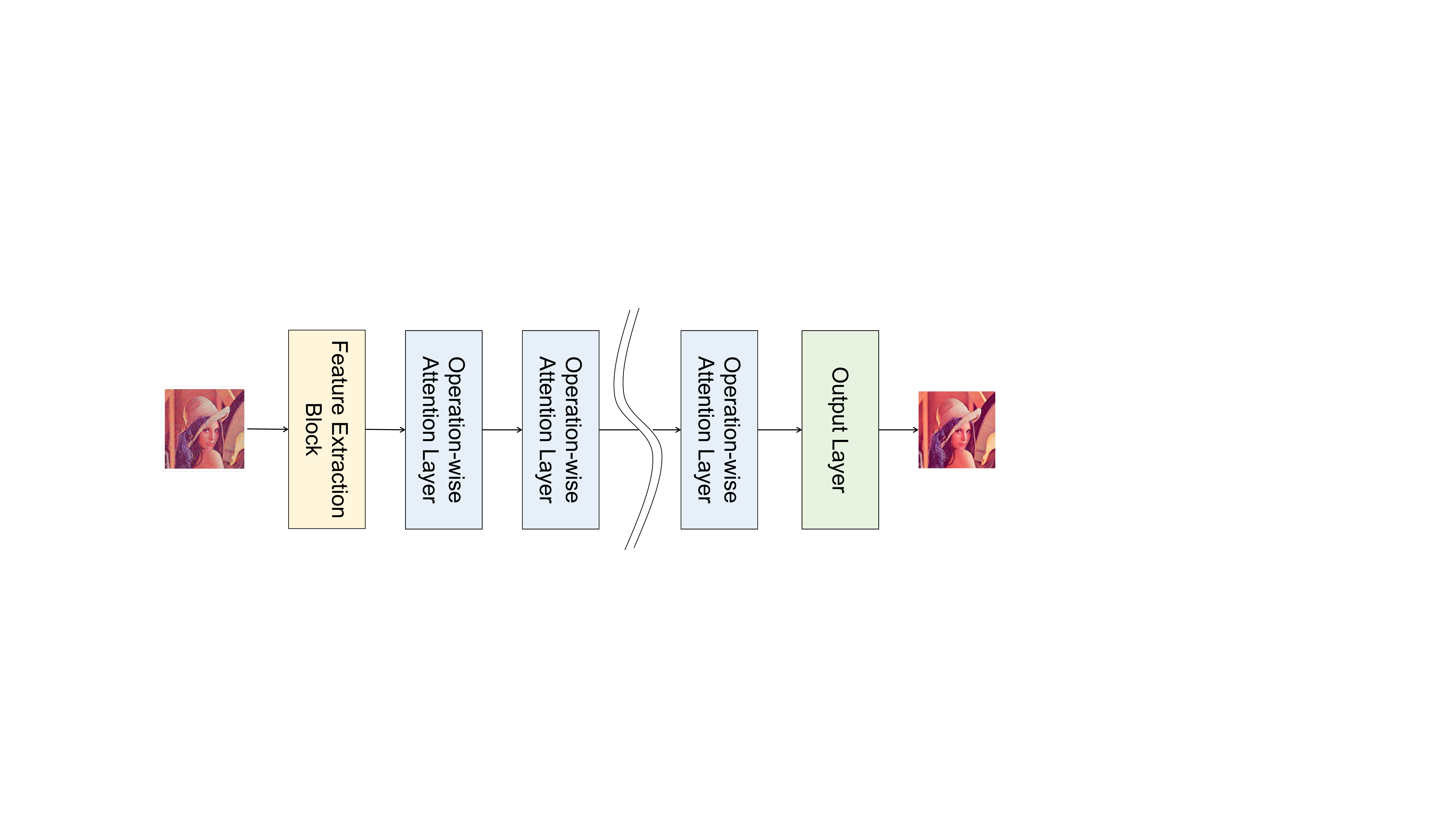}
\end{center}
   \caption{Overview of the operation-wise attention network. It consists of a feature extraction block, a stack of operation-wise attention layers, and an output layer.}
\label{fig:global}
\end{figure}

In this section, we describe the architecture of an entire network that employs the proposed operation-wise attention layers; see Fig.\ref{fig:global} for its overview.
It consists of three parts: a feature extraction block, a stack of operation-wise attention layers, and an output layer.
We first describe the operation-wise attention layer (Sec.\ref{owan}) and then explain the feature extraction block and the output layer (Sec.\ref{fe}).

\subsection{Operation-wise Attention Layer} \label{owan}

\subsubsection{Overview}
The operation-wise attention layer consists of an operation layer and an attention layer; see Fig.\ref{fig:owal}.
The operation layer contains multiple parallel operations, such as convolution and average pooling with different parameters.
The attention layer takes the feature map generated by the previous layer as inputs and computes attention weights on the parallel outputs of the operation layer. The operation outputs are multiplied with their attention weights and then concatenated to form the output of this layer. 
We intend this attention mechanism to work as a selector of the operations depending on the input.

\subsubsection{Operation-wise Attention}
We denote the output of the $l$-th operation-wise attention layer by ${\rm \bf{x}}_{\it l}$ $\in \mathbb{R}^{H\times W\times C}$ ($l=1,\ldots$), where $H$, $W$, and $C$ are its height, width, and the number of channels, respectively.
The input to the first layer in the stack of operation-wise attention layers, denoted by $\rm{\bf{x}}_0$, is the output of the feature extraction block connecting to the stack.
Let $\mathcal{O}$ be a set of operations contained in the operation layer; we use the same set for any layer $l$.  Given $\rm{\bf{x}}_{\it{l}-\rm{1}}$, we  calculate 
the attended value $\bar{a}_l^{o}$ on an operation $o(\cdot)$ in $\mathcal{O}$ as
\begin{eqnarray}
\bar{a}_l^o = \frac{\exp(\mathcal{F}_{l}({\rm \bf x}_{l-1}))}{\sum \exp(\mathcal{F}_{l}({\rm \bf x}_{l-1}))} = \frac{\exp(a_l^o)}{\sum_{o=1}^{\mathcal{|O|}} \exp(a_l^o)},
\label{attention}
\end{eqnarray}
where 
$\mathcal{F}_{l}$ is a mapping realized by the attention layer, which is given by
\begin{equation}
\mathcal{F}_{l}(
{\bf x}
) = \rm{\bf{W}}_2\sigma(\rm{\bf{W}}_1 \rm{\bf{z}}),
\label{attentionmap}
\end{equation}
where $\rm{\bf{W}}_1$ $\in \mathbb{R}^{T\times C}$ and $\rm{\bf{W}}_2$ $\in \mathbb{R}^{\mathcal{\mathcal{|O|}} \times T}$ are learnable weight matrices;
$\sigma(\cdot)$ denotes a ReLU function; and ${\bf{z}}\in \mathbb{R}^{C}$ is a vector containing the channel-wise averages of the input ${\bf{x}}$ as
\begin{equation}
z_c = \frac{1}{H\times W} \sum_{i=1}^{H}\sum_{j=1}^{W}x_{i,j,c}.
\end{equation}
Thus, we use the channel-wise average $\rm{\bf{z}}$ to generate attention weights ${\bf a}_l=[\bar{a}_l^{1},\ldots,\bar{a}_l^{\vert\mathcal O\vert}]$ instead of using full feature maps, which  is computationally expensive. 

\begin{figure*}[t]
\begin{center}
\includegraphics[scale=0.125]{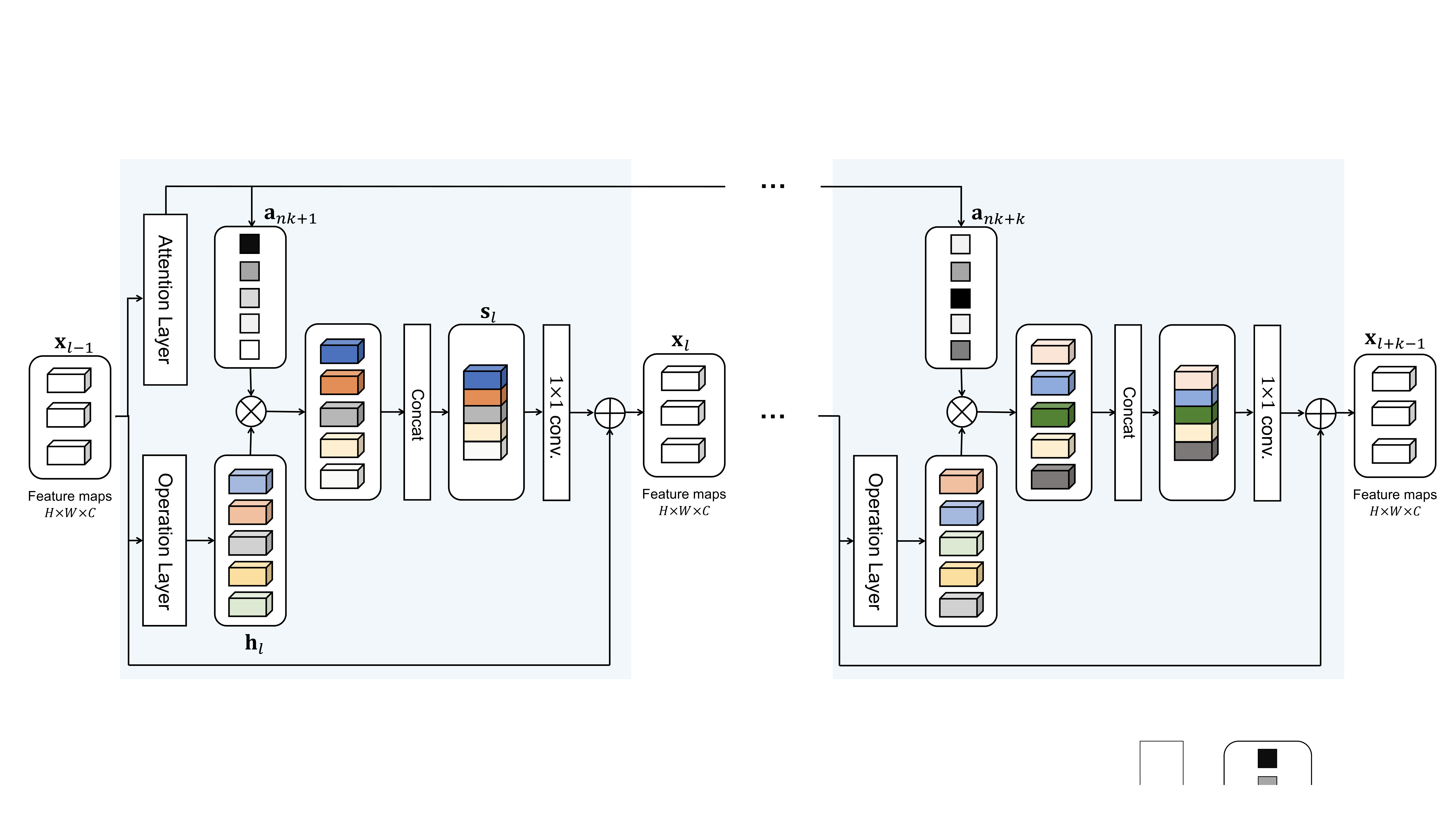}
\end{center}
   \caption{Architecture of the operation-wise attention layer. It consists of an attention layer, an operation layer, a concatenation operation, and $1\times 1$ convolution. Attention weights over operations of each layer are generated at the first layer in a group of consecutive $k$ layers. Note that different attention weights are generated for each layer.}
\label{fig:owal}
\end{figure*}

We found in our preliminary experiments that it makes training more stable to generate attention weights in the first layer of every few layers rather than to generate and use attention weights within each individual layer. (By layer, we mean operation-wise attention layer here.) 
To be specific, we compute the attention weights to be used in a group of $k$ consecutive layers at the first layer of the group; see Fig.\ref{fig:owal}. Letting $l=nk+1$ for a non-negative integer $n$, we compute attention weights ${\bf a}_{nk+1},\ldots, {\bf a}_{nk+k}$ at the $l$-th layer, where the computation of Eq.(\ref{attentionmap}) is performed using different ${\bf W}_1$ and ${\bf W}_2$ for each of ${\bf a}_{nk+1},\ldots, {\bf a}_{nk+k}$  but the same ${\bf x}$ and ${\bf z}$ of the $l$-th layer.
We will refer to this attention computation as {\em group attention}.

We multiply the outputs of the multiple operations with the attention weights computed as above. Let $f_o$ be the $o$-th operation and ${\bf h}_l^o(\equiv f_o({\bf x}_{l-1}))\in \mathbb{R}^{\rm{\it{H}}\times \it{W}\times \it{C}}$ be its output for $o=1,\ldots,\mathcal |O|$. We multiply ${\bf h}_l^o$'s with the attention weights $\bar{a}_l^o$'s, and then concatenate them in the channel dimension, obtaining ${\bf s}_l\in \mathbb{R}^{\rm{\it{H}}\times \it{W}\times \it{C}|\mathcal{O}|}$:
\begin{eqnarray}
{\rm \bf{s}}_{l} = {\rm Concat}[\bar{a}^1_l{\bf h}^1_l, \dots, \bar{a}^{|\mathcal{O}|}_l {\bf h}^{|\mathcal{O}|}_l].
\end{eqnarray}
The output of the $l$-th operation-wise attention layer is calculated by
\begin{eqnarray}
\rm{\bf{x}}_{\it{l}} = \mathcal{F}_{c}(\rm{\bf{s}}_{\it{l}}) + \rm{\bf{x}}_{\it{l}\rm{-1}},
\label{skip}
\end{eqnarray}
where $\mathcal{F}_{c}$ denotes a $1\times 1$ convolution operation with $C$ filters.
This operation makes activation of different channels interact with each other and adjusts the number of channels. We employ a skip connection between the input and the output of each operation-wise attention layer, as shown in Fig.~\ref{fig:owal}.

\begin{figure}[t]
\begin{center}
\includegraphics[scale=0.08]{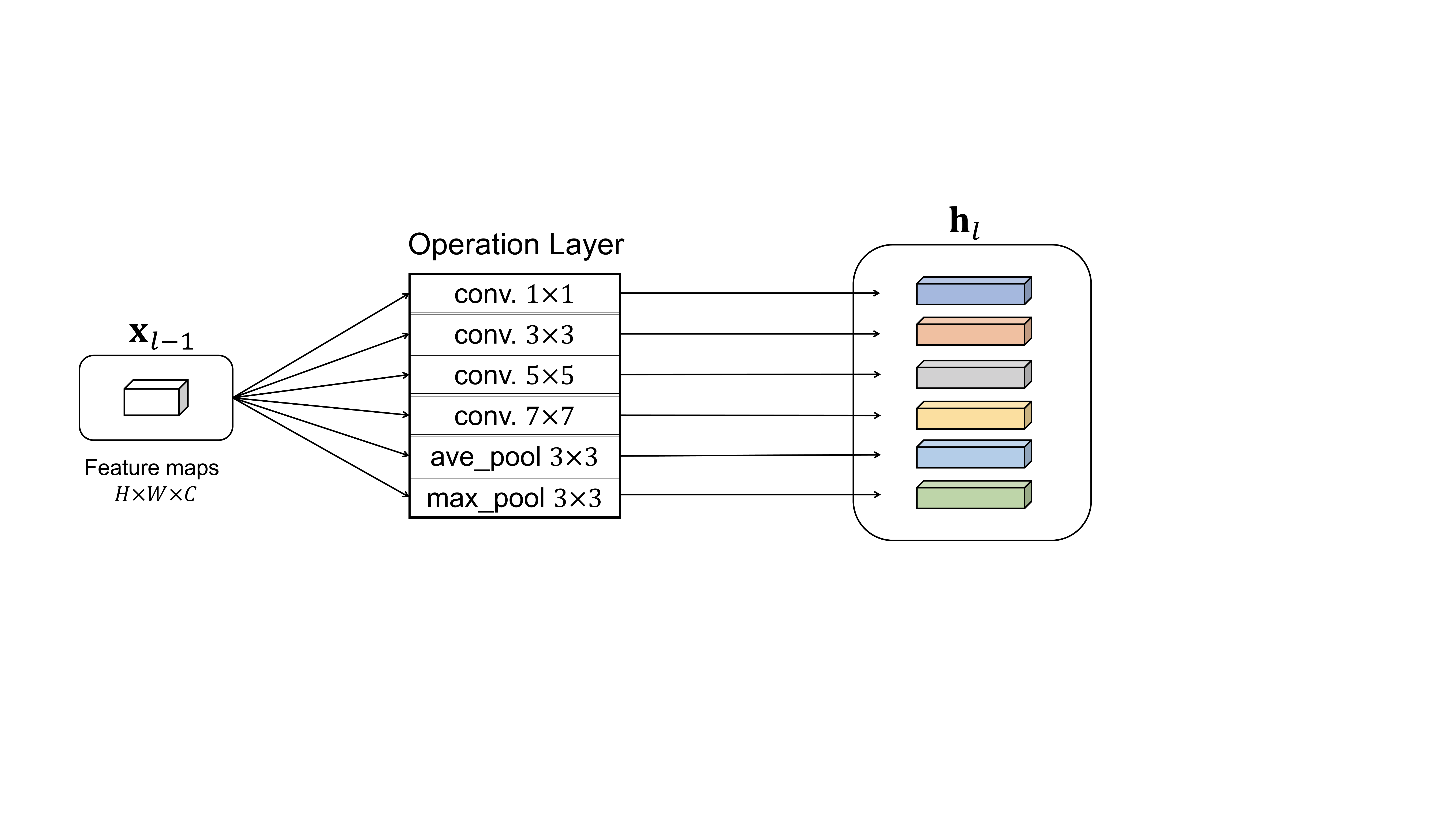}
\end{center}
\caption{An example of the operation layer in the operation-wise attention layer.}
\label{fig:op}
\end{figure}

\subsubsection{Operation Layer}

Considering the design of recent successful CNN models, 
we select $8$ popular operations for the operation layer: separable convolutions \cite{chollet2017xception} with filter sizes $1\times 1, 3\times 3, 5\times 5, 7\times 7$, dilated separable convolutions with filter sizes $3\times 3, 5\times 5, 7\times 7$ all with dilation rate $=2$, and average pooling with a $3\times 3$ receptive field.
All convolution operations use $C=16$ filters with stride $=1$, which is followed by a ReLU.
Also, we 
zero-pad the input feature maps computed in each operation not to change the sizes of its input and output.
As shown in Fig.\ref{fig:op}, the operations are performed in parallel, and they are concatenated in the channel dimension as mentioned above.

\subsection{Feature Extraction Block and Output Layer} \label{fe}

As mentioned earlier, our network consists of three parts, the feature extraction block, the stack of operation-wise attention layers, and the output layer. For the feature extraction block, we use a stack of standard residual blocks, specifically, $K$ residual blocks ($K=4$ in our experiments), in which each residual block has two convolution layers with $16$ filters of size $3\times 3$ followed by a ReLU. This block extracts features from a (distorted) input image and passes them to the operation-wise attention layer stack. 
For the output layer, we use a single convolution layer with kernel size $3\times 3$. The number of filters (i.e., output channels) is one if the input/output is a gray-scale image and three if it is a color image.

\begin{table*}[!t]
\centering
\caption{\bf{Results on DIV2K.} \rm{Comparison of DnCNN, RL-Restore, and our operation-wise attention network using DIV2K test sets. RL-Restore* displays the PSNR and SSIM values reported in \cite{Yu_2018_CVPR}.}}
\vskip 0.1in
\label{result1}
  \begin{tabular}{c|cc|cc|cc} \hline
    Test set & \multicolumn{2}{c|}{Mild (unseen)} & \multicolumn{2}{c|}{Moderate} & \multicolumn{2}{c}{Severe (unseen)} \\ \hline
    Metric & PSNR & SSIM & PSNR & SSIM & PSNR & SSIM \\ \hline
    DnCNN \cite{zhang_beyond_2017} & $27.51$ & $0.7315$ & $26.50$ & $0.6650$ & $25.26$ & $0.5974$ \\
    RL-Restore* \cite{Yu_2018_CVPR} & $28.04$ & $0.6498$ & $26.45$ & $0.5544$ & $25.20$ & $0.4629$ \\
    RL-Restore \cite{Yu_2018_CVPR} & $28.04$ & $0.7313$ & $26.45$ & $0.6557$ & $25.20$ & $0.5915$ \\
    Ours & $\bf{28.33}$ & $\bf{0.7455}$ & $\bf{27.07}$ & $\bf{0.6787}$ & $\bf{25.88}$ & $\bf{0.6167}$ \\ \hline
  \end{tabular}
\end{table*}

\begin{figure*}[!t]
\begin{center}
\includegraphics[scale=1.0]{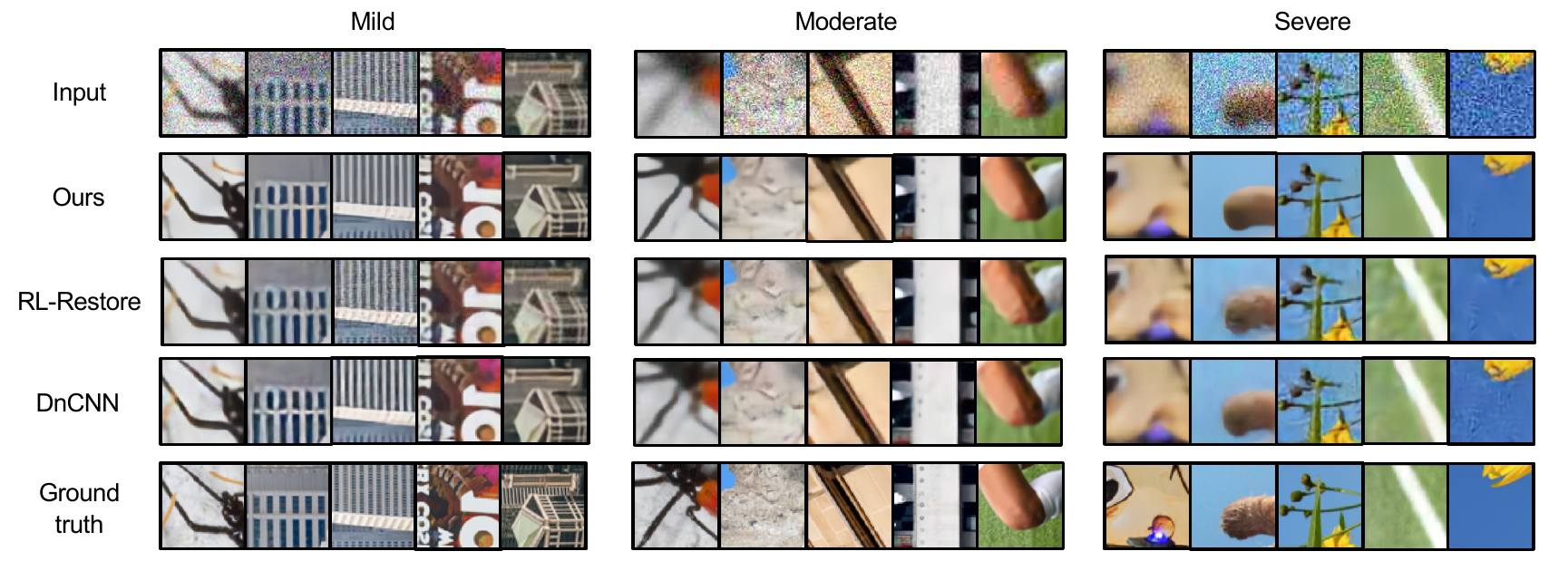}
\end{center}
\caption{Examples of restored images by our method, RL-Restore \cite{Yu_2018_CVPR}, and DnCNN \cite{zhang_beyond_2017}.}
\label{fig:mix}
\end{figure*}

\section{Experiments}

We conducted several experiments to evaluate the proposed method. 

\subsection{Experimental Configuration}

We used a network with $40$  operation-wise attention layers in all the experiments. We set the dimension of the weight matrices $\rm{\bf{W}}_1$, $\rm{\bf{W}}_2$ in each layer to $T=32$ and use 16 convolution filters in all the convolutional layers.
For the proposed group attention, we treat four consecutive operation-wise attention layers as a group (i.e., $k=4$). 

For the training loss, we use $\ell_1$ loss between the restored images and their ground truths as 
\begin{eqnarray}
L = \frac{1}{N}\sum_{n=1}^N \|OWAN({\bf y}_n) - {\bf x}_n\|_1,
\end{eqnarray}
where ${\rm \bf x}$ is a clean image, ${\rm \bf y}$ is its corrupted version, $N$ is the number of training samples, and $OWAN$ indicates the proposed operation-wise attention network.

In our experiments, 
we employed the Adam optimizer \cite{adam} with parameters $\alpha=0.001$, $\beta_1 = 0.9$, and $\beta_2 = 0.99$, along with  the cosine annealing technique of \cite{loshchilov2016sgdr} for adjusting the learning rate.
We trained our model for $100$ epochs with mini-batch size $32$.
We conducted all the experiments using PyTorch \cite{paszke2017automatic}.

\subsection{Performance on the Standard Dataset}

\subsubsection{DIV2K Dataset}

In the first experiment on combined distortions, we follow the experimental procedure described in \cite{Yu_2018_CVPR}.
They use the DIV2K dataset containing $800$ high-quality, large-scale images.
The $800$ images are divided into two parts: (1) the first $750$ images for training and (2) the remaining $50$ images for testing.
Then $63\times 63$ pixel patches are cropped from these images, yielding a training set and a testing set consisting of $249,344$ and $3,584$ patches, respectively.

They then apply multiple types of distortion to these patches.
Specifically, a sequence of Gaussian blur, Gaussian noise and JPEG compression is added to the training and testing images with different degradation levels.
The standard deviations of Gaussian blur and Gaussian noise are randomly chosen from the range of $[0, 5]$ and $[0, 50]$, respectively. The quality of JPEG compression is randomly chosen from the range of $[10, 100]$.
The resulting images are divided into three categories based on the applied degradation levels; mild, moderate, and severe (examples of the images are shown at the first row of Fig.\ref{fig:mix}).
The training are performed using only images of the moderate class, and testing are conducted on all three classes.

\begin{table*}[t]
\centering
\caption{\bf{Results on PASCAL VOC.} {\rm Comparison of RL-Restore \cite{Yu_2018_CVPR} and our method. A pretrained SSD300 \cite{liu2016ssd} is applied to distorted images (``w/o restoration'') and their restored versions.}}
\vskip 0.1in
\label{object}
\scalebox{0.61}{
  \begin{tabular}{c|c|c|cccccccccccccccccccc}
    VOC  & Method   & mAP  & aero & bike & bird & boat & bottle & bus & car & cat  & chair & cow & table& dog & horse & mbike&person& plant& sheep& sofa & train& tv \\ \hline
          &w/o restoration&$39.1$&$42.4$&$45.7$&$29.5$&$26.0$&$36.3$&$63.9$&$48.1$&$33.6$&$27.9$&$40.6$&$32.5$&$43.9$&$52.1$&$41.6$&$44.2$&$18.3$&$34.6$&$40.8$&$55.1$&$26.0$ \\
    2007  &RL-Restore \cite{Yu_2018_CVPR}&$66.0$&$74.9$&$78.2$&$60.1$&$\bf{43.2}$&$\bf{49.9}$&$\bf{80.8}$&$72.9$&$68.1$&$51.5$&$\bf{69.9}$&$63.4$&$76.2$&$78.6$&$82.6$&$68.3$&$\bf{41.6}$&$50.0$&$69.1$&$75.7$&$65.8$ \\
          &Ours&$\bf{69.3}$&$\bf{78.7}$&$\bf{78.3}$&$\bf{64.8}$&$41.2$&$47.8$&$\bf{80.8}$&$\bf{74.7}$&$\bf{69.4}$&$\bf{62.1}$&$68.7$&$\bf{74.7}$&$\bf{76.5}$&$\bf{79.2}$&$\bf{86.0}$&$\bf{69.0}$&$40.8$&$\bf{61.3}$&$\bf{80.4}$&$\bf{76.1}$&$\bf{76.6}$ \\ \hline
          &w/o restoration&$37.0$&$39.8$&$50.0$&$36.7$&$25.0$&$32.9$&$57.4$&$36.4$&$42.2$&$28.3$&$39.3$&$30.0$&$41.7$&$44.7$&$34.7$&$43.4$&$23.0$&$33.2$&$25.8$&$49.2$&$28.2$ \\
    2012  &RL-Restore \cite{Yu_2018_CVPR}&$62.0$&$73.4$&$73.8$&$64.0$&$41.2$&$50.5$&$76.7$&$59.3$&$69.8$&$45.8$&$58.8$&$59.9$&$72.2$&$72.3$&$76.6$&$69.3$&$41.3$&$51.0$&$53.0$&$71.5$&$59.6$ \\
         &Ours&$\bf{67.3}$&$\bf{77.6}$&$\bf{76.3}$&$\bf{69.2}$&$\bf{42.4}$&$\bf{52.0}$&$\bf{78.2}$&$\bf{64.5}$&$\bf{76.6}$&$\bf{56.6}$&$\bf{61.9}$&$\bf{70.8}$&$\bf{74.6}$&$\bf{76.0}$&$\bf{81.4}$&$\bf{71.0}$&$\bf{43.2}$&$\bf{57.7}$&$\bf{69.3}$&$\bf{72.1}$&$\bf{75.7}$ \\
  \end{tabular}
}
\end{table*}

\subsubsection{Comparison with State-of-the-art Methods}

We evaluate the performance of our method on the DIV2K dataset and compared with previous methods. 
In \cite{Yu_2018_CVPR}, the authors compared the performances of DnCNN \cite{zhang_beyond_2017} and their proposed method, RL-Restore, using PSNR and SSIM metrics. However, we found that the SSIM values computed in our experiments for these methods tend to be higher than those reported in \cite{Yu_2018_CVPR}, supposedly because of some difference in algorithm for calculating SSIM values. There are practically no difference in PSNR values.
For fair comparisons, we report here PSNR and SSIM values we calculated using the same code for all the methods. 
To run RL-Restore and DnCNN, we used the author's code for each\footnote{ https://github.com/yuke93/RL-Restore for RL-Restore and https://github.com/cszn/DnCNN/tree/master/TrainingCodes/\\dncnn\_pytorch for DnCNN.}.

Table \ref{result1} shows the PSNR and SSIM values of the three methods on different degradation levels of DIV2K test sets.
RL-Restore* indicates the numbers reported in \cite{Yu_2018_CVPR}.
It is seen that our method outperforms the two methods in all of the degradation levels in both the PSNR and SSIM metrics.

It should be noted that RL-Restore requires a set of pretrained CNNs for multiple expected distortion types and levels in advance; specifically, they train $12$ CNN models, where each model is trained on images with a single distortion type and level, e.g., Gaussian blur, Gaussian noise, and JPEG compression with a certain degradation level.
Our method does not require such pretrained models; it performs a standard training on a single model in an end-to-end manner.
This may be advantageous in application to real-world distorted images, since it is hard to identify in advance what types of distortion they undergo. 

We show examples of restored images obtained by our method, RL-Restore, and DnCNN along with their input images and ground truths in Fig.\ref{fig:mix}. They agree with the above observation on the PSNR/SSIM values shown in Table \ref{result1}.

\begin{figure}[t]
\begin{center}
\includegraphics[scale=0.33]{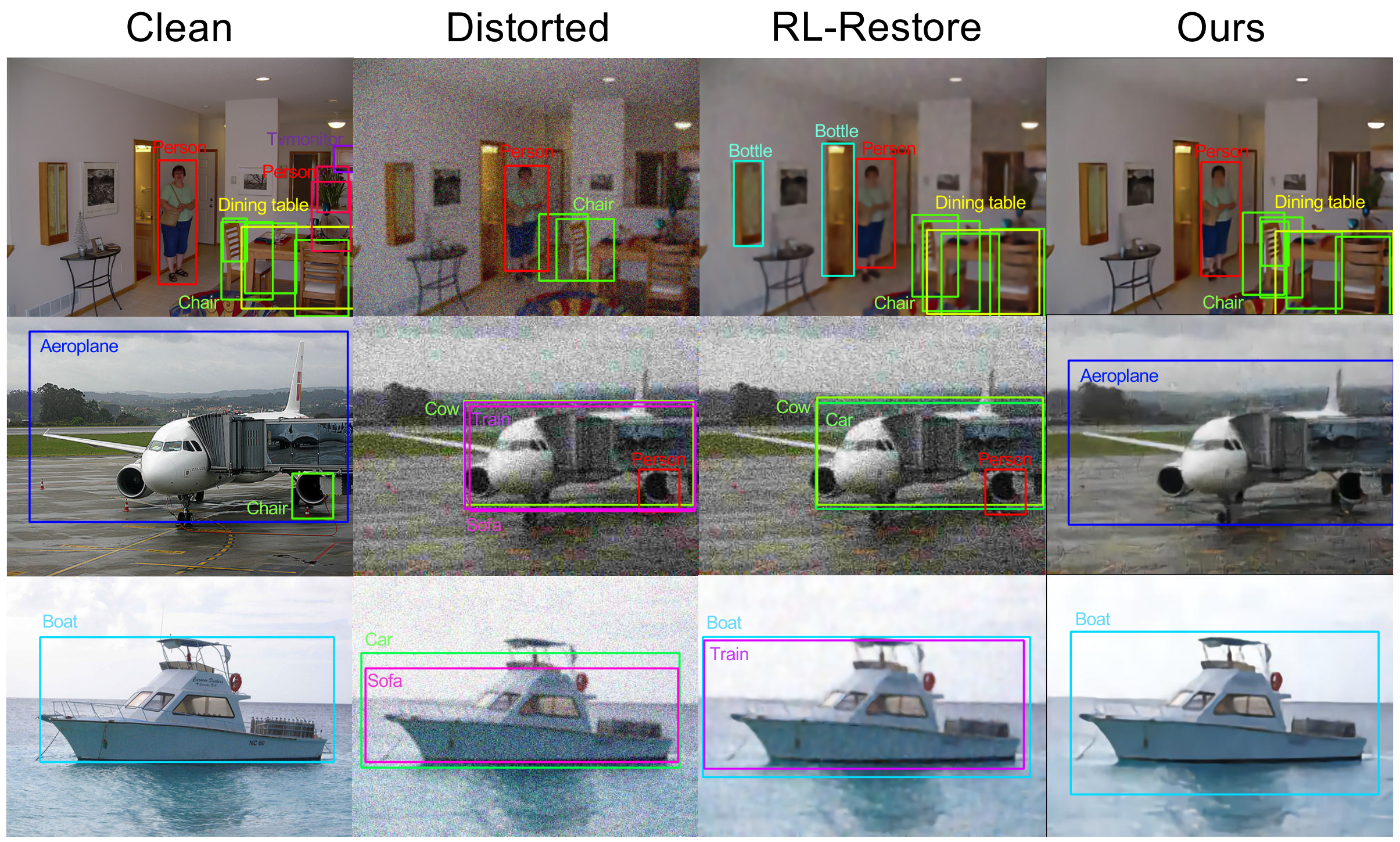}
\end{center}
\caption{Examples of results of object detection on PASCAL VOC. The box colors indicate class categories. }
\label{fig:object}
\end{figure}

\subsection{Evaluation on Object Detection}

To further compare the performance of our method and RL-Restore, we use the task of object detection. To be specific, we synthesized distorted images, then restored them and finally applied a pretrained object detector (we used SSD300 \cite{liu2016ssd} here) to the restored images.  
We used the images from the PASCAL VOC detection dataset. We added Gaussian blur, Gaussian Noise, and JPEG compression 
to the images of the validation set.
For each image, we randomly choose the degradation levels (i.e., standard deviations) for Gaussian blur and Gaussian noises from uniform distributions in the range $[0, 10]$ and $[30, 60]$, respectively, and the quality of JPEG compression from a uniform distribution with $[10, 50]$.

Table \ref{object} shows the mAP values obtained on (the distorted version of) the validation sets of PASCAL VOC2007 and VOC2012.
As we can see, our method improves detection accuracy by a large margin (around $30 \%$ mAP) compared to the distorted images.
Our method also provides better mAP results than RL-Restore for almost all categories. Figure \ref{fig:object} shows a few examples of detection results. It can be seen that our method removes the combined noises effectively, which we think contributes to the improvement of detection accuracy.

\subsection{Ablation Study} \label{sec:ab}

The key ingredient of the proposed method is the attention mechanism in the operation-wise attention layer. We conduct an ablation test to evaluate how it contributes to the performance. 

\subsubsection{Datasets} 

In this test, we constructed and used a different dataset of images with combined distortions.
We use the Raindrop dataset \cite{Qian_2018_CVPR} for the base image set.
Its training set contains $861$ pairs of a clean image of a scene and its distorted version with raindrops, and the test set contains $58$ images.
We first cropped $128\times 128$ pixel patches containing raindrops from each image and then added Gaussian noise, JPEG compression, and motion blur to them. 
We randomly changed the number of distortion types to be added to each patch, and also
randomly chose the level (i.e., the standard deviation) of the Gaussian noise and the quality of JPEG compression from the range of $[10, 30]$ and $[15, 35]$, respectively.
For motion blur, we followed the method of random trajectories generation of \cite{boracchi2012modeling}.
To generate several levels of motion blur, we randomly sampled the max length of trajectories from $[10, 80]$; see \cite{boracchi2012modeling, Kupyn_2018_CVPR} for the details.
As a result, the training and test set contain $50,000$ and $5,000$ 
patches, respectively.
We additionally created four sets of images with a single distortion type, 
each of which consists of $1,000$ patches with the same size. We used the same procedure of randomly choosing the degradation levels for each distortion type.
Note that we do not use the four single-distortion-type datasets for training.

\begin{table*}[!t]
\centering
\caption{\bf{Effects of the proposed attention mechanism.} \rm{Comparison with two baseline methods ({\it w/o attention} and {\it fixed attention}) on test sets of the new dataset we created. Each model is trained on a single set of images with combined distortions and evaluated on different sets of images with combined or single-type distortion.}}
\vskip 0.1in
\label{ablation}
  \begin{tabular}{c|cc|cc|cc|cc|cc} \hline
    Test set & \multicolumn{2}{c|}{Mix} & \multicolumn{2}{c|}{Raindrop} & \multicolumn{2}{c|}{Blur} & \multicolumn{2}{c|}{Noise} & \multicolumn{2}{c}{JPEG} \\ \hline
    Metric & PSNR & SSIM & PSNR & SSIM & PSNR & SSIM & PSNR & SSIM & PSNR & SSIM \\ \hline
    w/o attention & $23.24$ & $0.7342$ & $26.93$ & $0.8393$ & $21.74$ & $0.7546$ & $29.88$ & $0.8771$ & $29.09$ & $0.8565$ \\
    fixed attention & $23.37$ & $0.7345$ & $26.83$ & $0.8433$ & $21.99$ & $0.7590$ & $30.58$ & $0.8907$ & $29.20$ & $0.8609$ \\
    Ours & $\bf{24.71}$ & $\bf{0.7933}$ & $\bf{28.24}$ & $\bf{0.8764}$ & $\bf{23.66}$ & $\bf{0.8211}$ & $\bf{31.93}$ & $\bf{0.9102}$ & $\bf{30.07}$ & $\bf{0.8779}$ \\ \hline
  \end{tabular}
\end{table*}

\begin{figure*}[t]
\begin{center}
\includegraphics[scale=0.74]{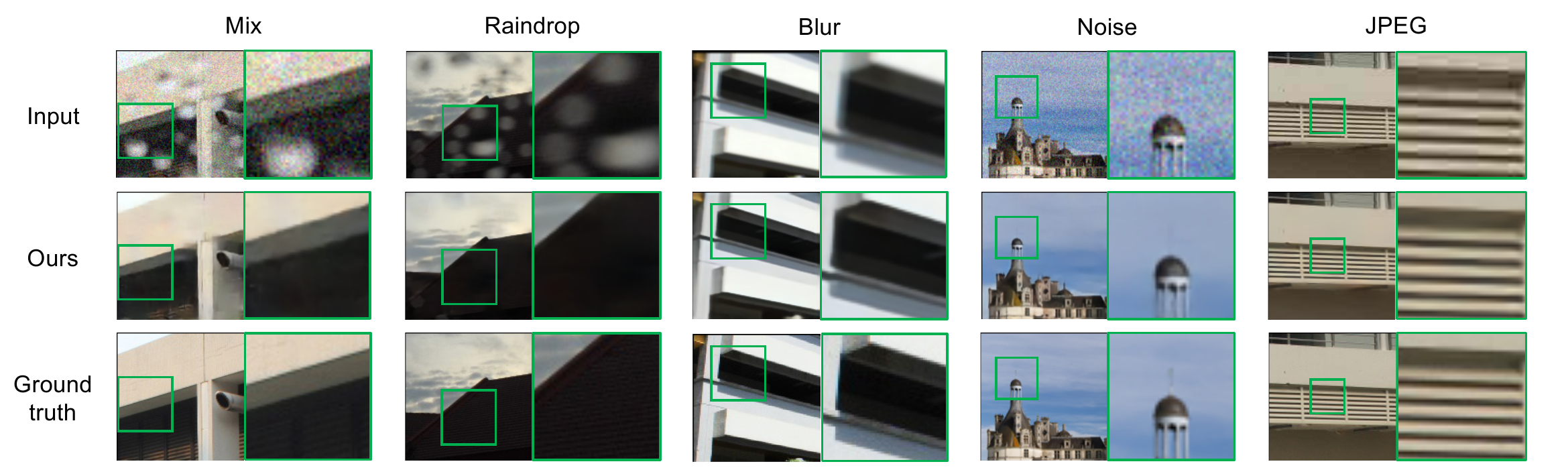}
\end{center}
\caption{Examples of restored images by our method from distorted images with a different type of distortion. The model is trained on a single set of images with combined distortions.}
\label{fig:mydata}
\end{figure*}

\subsubsection{Baseline Methods}

We consider two baseline methods for comparison, which are obtained by invalidating some parts of the attention mechanism in our method. 
One is to remove the attention layer from the operation-wise attention layer. In each layer, the parallel outputs from its operation layer are simply concatenated in the channel dimension and inputted into the $1\times 1$ convolution layer. Others are kept the same as in the original network. In other words, we set all the attention weights to 1 in the proposed network. We refer to this configuration as ``{ w/o attention}''.

The other is to remove the attention layer but employ constant attention weights on the operations. The attention weights no longer depend on the input signal.
We determine them together with the convolutional kernel weights by gradient descent. More rigorously, we alternately update the kernel weights and attention weights as in  \cite{liu2018darts}, a method proposed as differentiable neural architecture search. 
These attention weights are shared in the group of $k$ layers; we set $k=4$ in the experiments as in the proposed method.
We refer to this model as `` fixed attention'' in what follows.
These two models along with the proposed model are trained on the dataset with combined distortions and then evaluated on each of the aforementioned datasets.

\subsubsection{Results}

The quantitative results are shown in Table \ref{ablation}.
It is observed that the proposed method works much better than the two baselines on all distortion types in terms of PSNR and SSIM values.
This confirms the effectiveness of the attention mechanism in the proposed method. We show several examples of images restored by the proposed method in Fig.\ref{fig:mydata}. It can be seen that not only images with combined distortions but those with a single type of distortion (i.e., raindrops, blur, noise, JPEG compression  artifacts) are recovered fairly accurately.  


\subsection{Analysis of Operation-wise Attention Weights}

\begin{figure}[t]
\begin{center}
\includegraphics[scale=0.6]{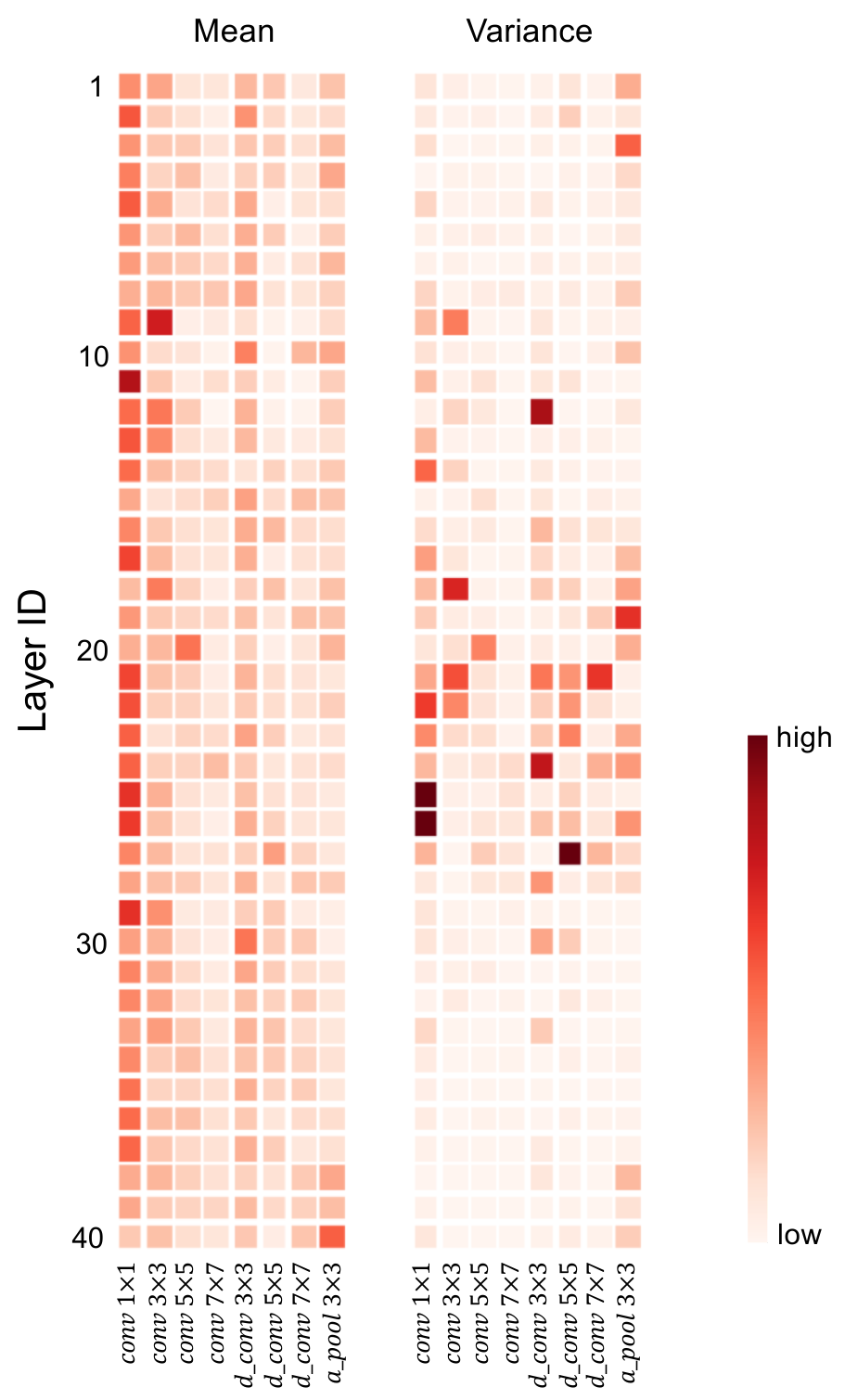}
\end{center}
\caption{Mean and variance of attention weights over the input images with a single type of distortion, i.e., raindrop, blur, noise, JPEG artifacts.}
\label{fig:attention}
\end{figure}

To analyze how the proposed operation-wise attention mechanism works, we visualize statistics of the attention weights of all the layers for the images with a single type of distortion.
Figure \ref{fig:attention} shows the mean and variance of the attention weights over the four single-distortion datasets, i.e., raindrops, blur, noise, and JPEG artifacts. 
Each row and column indicates one of the forty operation-wise attention layers and one of eight operations employed in each layer, respectively.

We can make the following observations from the map of mean attention weights (Fig.~\ref{fig:attention}, left): i) $1\times 1$ convolution tends to be more attended than the others throughout the layers;
and ii) convolution with larger-size filters, such as $5\times 5$ and $7\times 7$, tend to be less attended. 
It can be seen from the variance map (Fig.~\ref{fig:attention}, right) that 
iii) attention weights have higher variances for middle layers than for other layers, indicating the tendency that the middle layers more often change the selection of operations than other layers.

\begin{figure}[t]
\begin{center}
\includegraphics[scale=0.59]{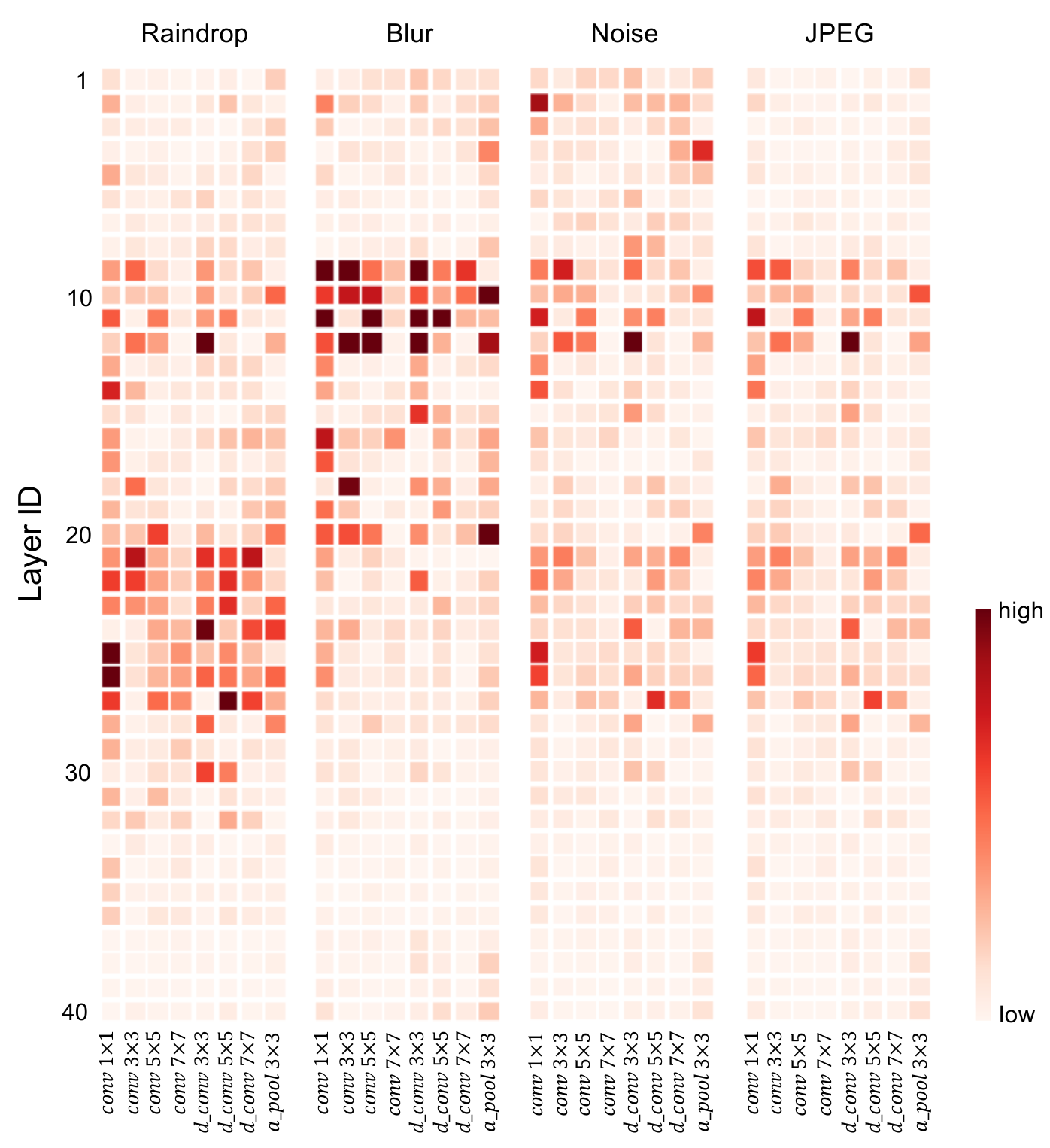}
\end{center}
\caption{Visualization of mean attention weights for four distortion types. Each element indicates the difference between the mean attention weights over each distortion-type images and those over all the images (the left map of Fig.\ref{fig:attention}).}
\label{fig:diff}
\end{figure}

Figure \ref{fig:diff} shows the absolute differences between the mean attention weights over each of four single-distortion datasets and the mean attention weights over all the four datasets.
It is observed that the attention weights differ depending on the distortion type of the input images. 
This indicates that the proposed attention mechanism does select operations depending on the distortion type(s) of the input image. 
It is also seen that the map for raindrop and that for blur are considerably different from each other and the other two, whereas those for noise and JPEG are mostly similar, although there are  some differences in the lower layers. This implies the (dis)similarity among the tasks dealing with these four types of distortion.

\subsection{Performance on novel strengths of distortion}
To test our method on a wider range of distortions, we evaluate its performance on novel strengths of distortion that have not been learned.
Specifically, we created a training set by randomly sampling the standard deviations of Gaussian noise and the quality of JPEG compression from $[0, 20]$ and $[60, 100]$, respectively and sampling the max length of trajectories for motion blur from $[10, 40]$.
We then apply trained models to a test set with a different range of distortion parameters; the test set is created by sampling the standard deviations of Gaussian noise, the quality of JPEG compression, and the max trajectory length from $[20, 40]$, $[15, 60]$, and $[40, 80]$, respectively.
We used the same image set used in the ablation study for the base image set.
The results are shown in Table \ref{range}, which shows that our method outperforms the baseline method by a large margin.

\begin{table}[t]
\centering
\caption{Performance on novel strengths of distortion.}
\vskip 0.1in
\label{range}
  \begin{tabular}{c|cc} \hline
    Metric & PSNR & SSIM \\ \hline
    DnCNN \cite{zhang_beyond_2017} & $18.94$ & $0.3594$  \\
    Ours & $\bf{23.07}$ & $\bf{0.6795}$ \\ \hline
  \end{tabular}
\end{table}

\section{Conclusion}

In this paper, we have presented a simple network architecture, named operation-wise attention mechanism, for the task of restoring images having combined distortions with unknown mixture ratios and strengths. It enables to 
attend on multiple operations performed in a layer depending on input signals.
We have proposed a layer with this attention mechanism, which  can be stacked to build a deep network. The network is differentiable and can be trained in an end-to-end fashion using gradient descent. The experimental results show that the proposed method works better than the previous methods on the tasks of image restoration with combined distortions, proving the effectiveness of the proposed method.

\section*{Acknowledgement}
This work was partly supported by JSPS KAKENHI Grant Number JP15H05919 and JST CREST Grant Number JPMJCR14D1.

{\small
\bibliographystyle{ieee}
\bibliography{camera}
}

\clearpage
\appendix

\setcounter{figure}{8}
\setcounter{table}{3}

\section{Comparison with Existing Methods Dedicated to Single Types of Distortion}
\label{sec1}

In this study, we consider restoration of images with combined distortion of multiple types. To further analyze effectiveness of the proposed method, we compare it with existing methods dedicated to single types of distortion on their target tasks, i.e.,  single-distortion image restoration.
To be specific, for the four single-distortion tasks, i.e.,  removal of Gaussian noise, motion blur, JPEG artifacts, and raindrops, we compare the proposed method with existing methods that are designed for each individual task and trained on the corresponding (single-distortion) dataset. On the other hand, the proposed model (the same as the one explained in the main paper) is trained on the combined distortion dataset explained in Sec.~4.4.1. 
Then, they are tested on the test splits of the same single-distortion datasets. 

As this setup is favorable for the dedicated methods, they are expected to yield better results. The purpose of this experiment is to understand how large the differences will be.

\subsection{Noise Removal}
\paragraph{Experimental Configuration}

We use the DIV2K dataset for the base image set.
We first cropped $128\times 128$ pixel patches from the training and validation sets of the DIV2K dataset.
Then we added Gaussian noise to them, yielding a training set and a testing set which consist of $50,000$ and $1,000$ patches, respectively.
The standard deviation of the Gaussian noise is randomly chosen from the range of $[10, 20]$.
Using these datasets, we compare our method with DnCNN \cite{zhang_beyond_2017}, FFDNet \cite{zhang2018ffdnet}, and E-CAE \cite{suganumaICML2018}, which are the state-of-the-art dedicated models for this task.

\begin{figure}[t]
\begin{center}
\includegraphics[scale=0.8]{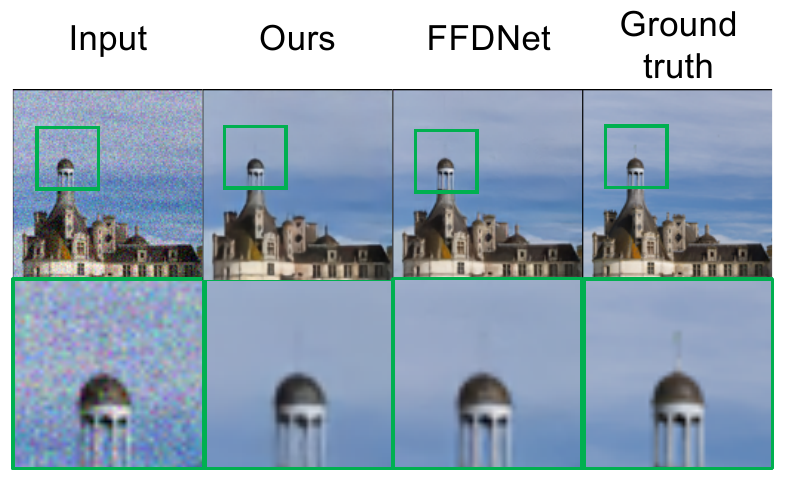}
\end{center}
  \caption{Examples of noise removal by our method and FFDNet \cite{zhang2018ffdnet}.}
\label{fig:noise}
\end{figure}

\paragraph{Results}
Table \ref{noise} shows the results. 
As expected, the proposed method is inferior to the state-of-the-art methods by a certain margin. Although the margin is quantitatively not small considering the differences among the state-of-the-art methods, qualitative differences are not so large; an example is shown in Fig.~\ref{fig:noise}. 

To evaluate the potential of the proposed method, we also report the performance of our model trained on the single-distortion training data, which is referred to as `Ours*' in the table.
It is seen that it achieves similar or even better accuracy. This proves the potential of the proposed model as well as implies its usefulness in the scenario where there is only a single but {\em unidentified} type of distortion in images for which training data are available.

\begin{table}[!t]
\centering
\caption{Results on the Gaussian noise removal.}
\vskip 0.1in
\label{noise}
  \begin{tabular}{c|cc} \hline
    Method & PSNR & SSIM \\ \hline
    DnCNN \cite{zhang_beyond_2017} & $34.38$ & $0.9289$ \\
    FFDNet \cite{zhang2018ffdnet}  & $34.90$ & $0.9355$ \\
    E-CAE \cite{suganumaICML2018}  & $35.08$ & $0.9365$ \\
    Ours                           & $31.49$ & $0.8972$ \\ 
    Ours*                          & $35.20$ & $0.9381$ \\ \hline
  \end{tabular}
\end{table}

\subsection{JPEG Artifacts Removal}

\paragraph{Experimental Configuration}
We follow the same procedure as noise removal to construct a training set and a test set. 
The quality of the JPEG compression is randomly chosen from the range of $[15, 35]$.
We compare our method with two single distortion methods, DnCNN \cite{zhang_beyond_2017} and MemNet \cite{mem}.

\begin{figure}[t]
\begin{center}
\includegraphics[scale=0.8]{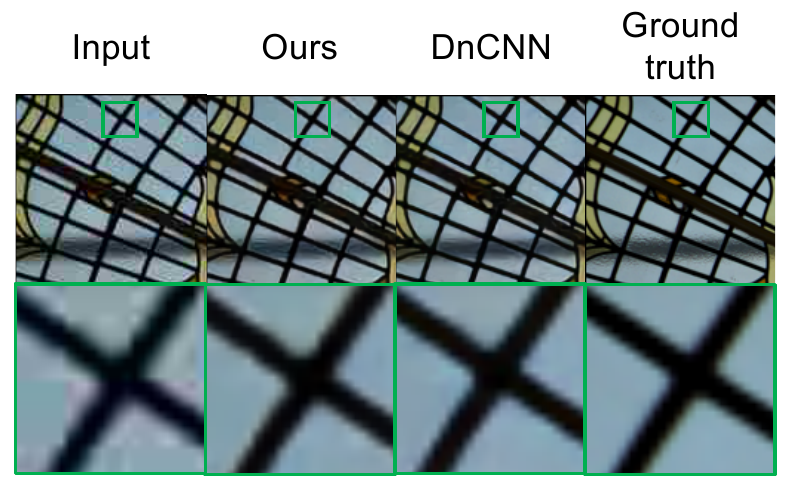}
\end{center}
  \caption{Examples of JPEG artifact removal by our method and DnCNN \cite{zhang_beyond_2017}.}
\label{fig:jpeg}
\end{figure}

\paragraph{Results}
Table \ref{JPEG} shows the results.
We can observe the same tendency as noise removal with smaller qualitative differences between our method and the dedicated methods.
Their qualitative gap is further smaller; examples of restored images are shown in 
Fig.\ref{fig:jpeg}. Our method trained on the same single distortion dataset (`Ours*') achieves better results with noticeable differences in this case. 

\begin{table}[!t]
\centering
\caption{Results of JPEG artifact removal.}
\vskip 0.1in
\label{JPEG}
  \begin{tabular}{c|cc} \hline
    Method & PSNR & SSIM \\ \hline
    DnCNN \cite{zhang_beyond_2017} & $31.24$ & $0.8827$ \\
    MemNet \cite{mem}              & $30.85$ & $0.8785$ \\
    Ours                           & $29.87$ & $0.8684$ \\
    Ours*                          & $31.64$ & $0.8902$ \\ \hline
  \end{tabular}
\end{table}

\begin{table}[!t]
\centering
\caption{Results of  motion blur removal.}
\vskip 0.1in
\label{blur}
  \begin{tabular}{c|cc} \hline
    Method & PSNR & SSIM \\ \hline
    Sun \etal \cite{sun2015learning}                     & $24.6$ & $0.842$ \\
    Nah \etal \cite{nah2017deep}                         & $29.1$ & $0.916$ \\
    Xu \etal \cite{xu2013unnatural}                      & $25.1$ & $0.890$ \\
    DeblurGAN ({\it Synth}) \cite{Kupyn_2018_CVPR}       & $23.6$ & $0.884$ \\
    DeblurGAN ({\it Wild}) \cite{Kupyn_2018_CVPR}        & $27.2$ & $0.954$ \\
    DeblurGAN ({\it Comb}) \cite{Kupyn_2018_CVPR}        & $28.7$ & $0.958$ \\
    Ours                                                 & $25.4$ & $0.793$ \\\hline
  \end{tabular}
\end{table}

\begin{figure}[t]
\begin{center}
\includegraphics[scale=0.35]{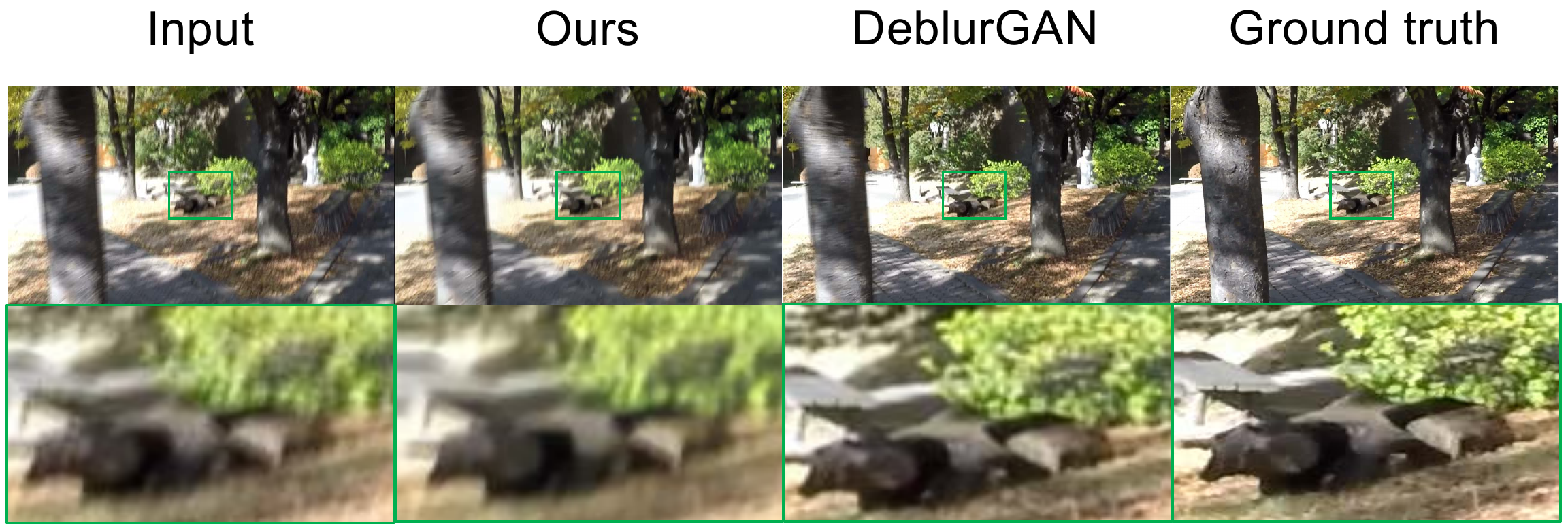}
\end{center}
  \caption{Examples of motion blur removal by our method and DeblurGAN ({\it Wild}) \cite{Kupyn_2018_CVPR}.}
\label{fig:blur}
\end{figure}

\begin{table}[!t]
\centering
\caption{Results on the raindrop removal.}
\vskip 0.1in
\label{raindrop}
  \begin{tabular}{c|cc} \hline
    Method & PSNR & SSIM \\ \hline
    Attentive GAN \cite{Qian_2018_CVPR}           & $31.57$ & $0.9023$ \\
    Attentive GAN ({\it w/o D}) \cite{Qian_2018_CVPR}    & $29.25$ & $0.7853$ \\
    Ours                                          & $28.57$ & $0.8878$ \\ \hline
  \end{tabular}
\end{table}

\begin{figure}[t]
\begin{center}
\includegraphics[scale=0.38]{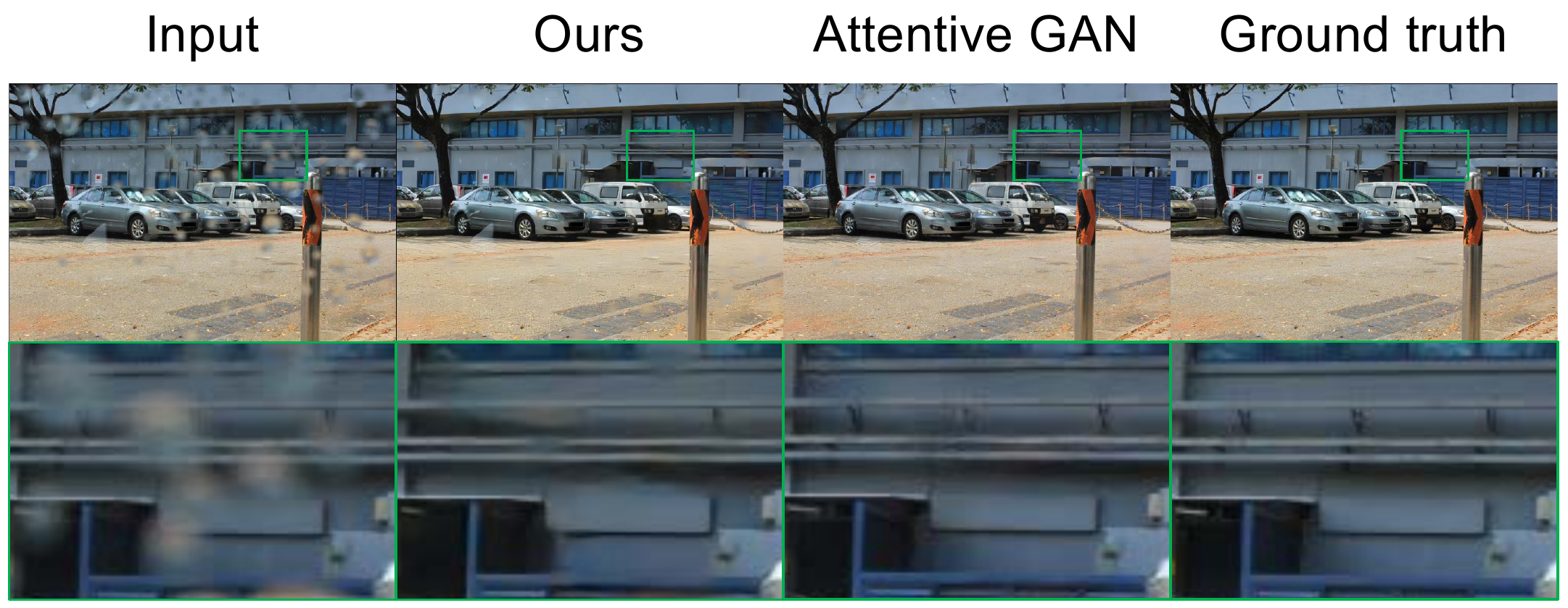}
\end{center}
  \caption{Examples of raindrop removal by our method and Attentive GAN \cite{Qian_2018_CVPR}.}
\label{fig:rain}
\end{figure}

\subsection{Blur Removal}

\paragraph{Experimental Configuration}

We used the  GoPro dataset \cite{nah2017deep} for evaluation. 
It consists of 2,013 training and 1,111 test pairs of blurred and sharp images. We compare our model, which is trained on the combined distortion dataset, to the state-of-the-art method of  \cite{Kupyn_2018_CVPR}.
In \cite{Kupyn_2018_CVPR}, the authors provide several versions of their method; DeblurGAN ({\it Synth}), DeblurGAN ({\it Wild}), and DeblurGAN ({\it Comb}).
DeblurGAN ({\it Synth}) indicates a version of their model trained on a synthetic dataset generated by the method of \cite{boracchi2012modeling}, which is also employed in our experiments.
Note that the dataset for training DeblurGAN ({\it Synth}) is generated from images of the MS COCO dataset, whereas the dataset for training our model is generated from the Raindrop dataset \cite{Qian_2018_CVPR} and moreover it contains combined distortions.
DeblurGAN ({\it Wild}) indicates a version of their model trained on random crops from the GoPro training set \cite{nah2017deep}.
DeblurGAN ({\it Comb}) is a model trained on the both datasets. 

\paragraph{Results}

Table \ref{blur} shows the results
for the three variants of DeblurGAN along with three other existing methods, where their PSNR and SSIM values are copied from \cite{Kupyn_2018_CVPR}.
It is seen that our method is comparable to the existing ones, in particular earlier methods, in terms of PSNR but is inferior in terms of SSIM by a large margin.
In fact, qualitative difference between our method and DeblurGAN ({\it Wild}) is large; see Fig.\ref{fig:blur}.

However, we think that some of the gap can be explained by the difference in training data. DeblurGAN ({\it Wild}) is trained using the dataset lying in the same domain as the test data. 
On the other hand, our model is trained only on synthetic data, which must have different distribution from the GoPro test set. Thus, it may be fairer to make qualitative comparison with DeblurGAN ({\it Synth}), but we do not do so here, since its implementation is unavailable.

\begin{figure*}[!ht]
\begin{center}
\includegraphics[scale=0.8]{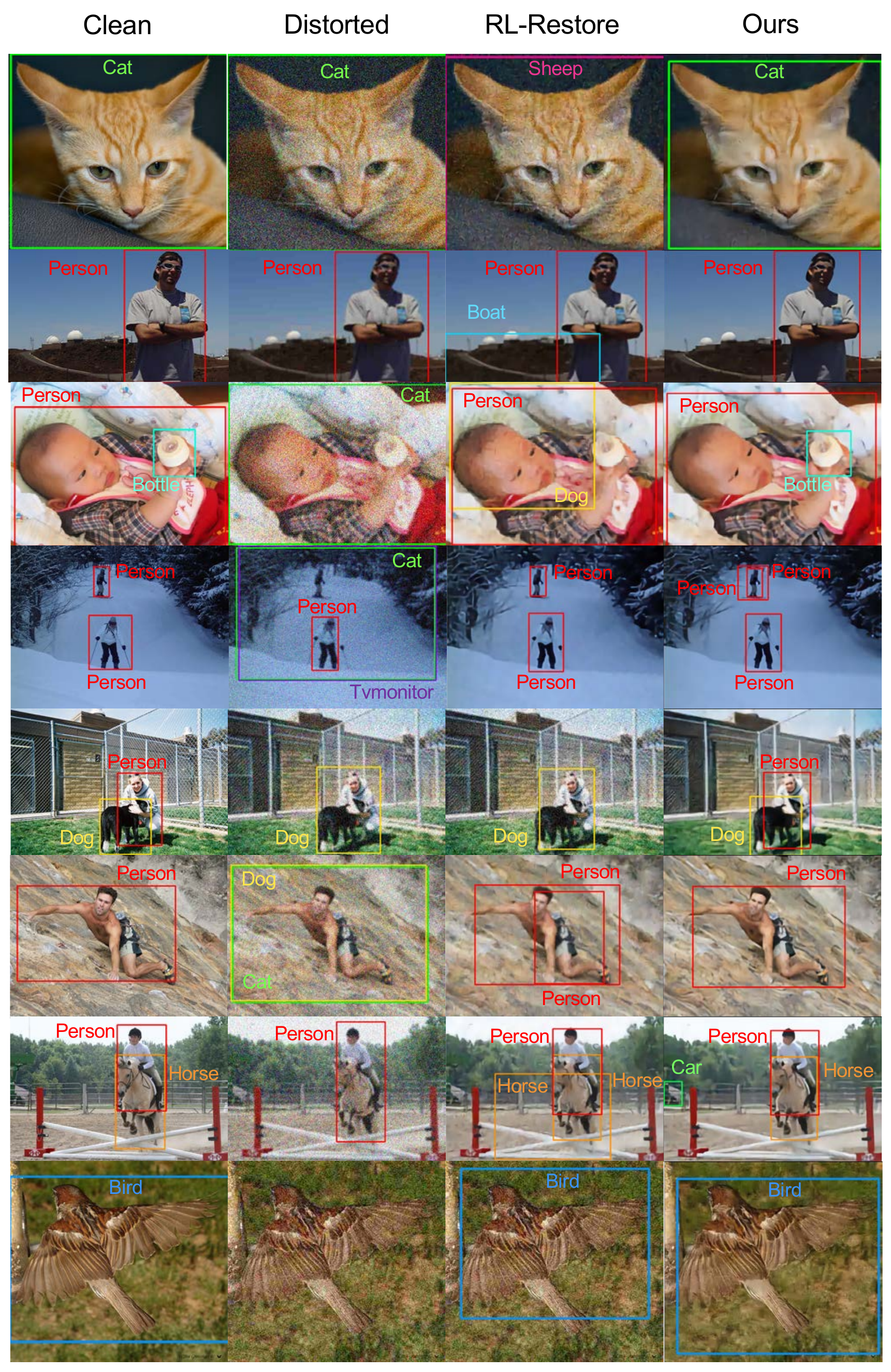}
\end{center}
  \caption{Examples of object detection results on PASCAL VOC. The box colors indicate class categories.}
\label{fig:detect}
\end{figure*}

\subsection{Raindrop Removal}
\paragraph{Experimental Configuration}

We use the Raindrop dataset \cite{Qian_2018_CVPR}, which contains two test sets called TestA and TestB; the former is a subset of the latter.
We use TestA for evaluation following \cite{Qian_2018_CVPR}.
We compare our model, which is trained on the combined distortion dataset as above, against the state-of-the-art method, Attentive GAN \cite{Qian_2018_CVPR}.
 
\paragraph{Results}
Table \ref{raindrop} shows the results. 
Attentive GAN ({\it w/o D}) is a variant that is trained without a discriminator and with only combined loss functions of the mean squared error and the perceptual loss etc. 
Our model achieves slightly lower accuracy than Attentive GAN ({\it w/o D}) in terms of PSNR and slightly lower accuracy than Attentive GAN in terms of SSIM. 
Figure \ref{fig:rain} shows example images obtained by our method and Attentive GAN. Although there is noticeable difference between the images generated by the two methods, it is fair to say that our method yields reasonably good result, considering the fact that {\em our model can handle other types of distortion as well. }

\section{More Results of Object Detection From Distorted Images}  \label{sec2}

We have shown a few examples of object detection on PASCAL VOC in Fig.5 of the main paper.
We provide more examples in Fig.\ref{fig:detect}.
They demonstrate that our method is able to remove combined distortions effectively, which contributes to the improvement of detection accuracy.

\end{document}